\documentclass[runningheads]{llncs}
\usepackage{graphicx}
\usepackage{listings}
\usepackage{hyperref} %for hyperlinks in the thesis
\usepackage[dvipsnames]{xcolor}
\hypersetup{
	colorlinks,
	linkcolor={red!50!black},
	citecolor={blue!50!black},
	urlcolor={blue!80!black}
}
\usepackage{tikz}
\usepackage{subcaption}
\newcommand*\nn[1]{\overline{#1}}

{\end{list}}

\newcounter{myenumctr}

\newcommand*{\overtabline}{%
	\noalign{%
		\vskip-.5\dimexpr\ht\@arstrutbox+\dp\@arstrutbox\relax
		\vskip-.2pt\relax
		\hrule
		\vskip-.2pt\relax
		\vskip+.5\dimexpr\ht\@arstrutbox+\dp\@arstrutbox\relax
	}%
}

\makeatletter
\renewcommand*{\p@section}{\S\,}
\renewcommand*{\p@subsection}{\S\,}
\renewcommand*{\p@subsubsection}{\S\,}
\makeatother

\begin{document}

\title{Logic Conditionals, Supervenience,\\and Selection Tasks %\thanks{Grants or other notes
}


\author{Giovanni Sileno\thanks{The author would like to thank Jean-Louis Dessalles, Isabelle Bloch and Alexander Boer for comments and suggestions provided to earlier versions of the paper. % This research was partly funded by NWO (VWData project), and by ANR (LOGIMA).
	}%\inst{1}\orcidID{0000-1111-2222-3333}%         \and
	%Second Author %etc.
}


\institute{Informatics Institute, University of Amsterdam, Netherlands \\
	\email{g.sileno@uva.nl}           %  \\
}

\maketitle              % typeset the header of the contribution

\vspace{-10pt}
\begin{abstract}
Principles of cognitive economy would require that concepts about objects, properties and relations should be introduced only if they simplify the conceptualisation of a domain. Unexpectedly, classic logic conditionals, specifying structures holding within elements of a formal conceptualisation, do not always satisfy this crucial principle. The paper argues that this requirement is captured by \emph{supervenience}, hereby further identified as a property necessary for compression. The resulting theory suggests an alternative explanation of the empirical experiences observable in selection tasks, associating human performance with conditionals on the ability of dealing with compression, rather than with logic necessity. 
\vspace{-5pt}
\end{abstract}

\vspace{-15pt}
\section{Introduction}
\vspace{-5pt}
The difficulties---if not the inadequacy---of formal logic in modelling human cognition have been claimed in the literature by numerous authors. % The content of their claims generally varies, from completely denying any role of logic in cognition (e.g. emphasising non-deterministic aspects of human reasoning), to proposing alternative accounts to classic or non-standard logics (e.g. non-monotonic logics) or geometric models of cognition, to take explicitly into consideration certain \emph{biases} by identifying representative heuristics. 
Within this discussion, the celebrity of Wason's selection task \cite{Wason1968} is on par with the simplicity of the experiment and the unexpectedness of the results. The wide presence of rule-like conceptual structures (usually in the form of conditionals \textit{if.. then..}) in formal and semi-formal structurations of knowledge highly contrasts with the picture of the human ability of dealing with rules captured by this family of experiments.

Many hypotheses have been proposed in the last 50 years to explain human performance in selection tasks, e.g. primitive \textit{matching bias} \cite{Evans2008}, the influence of \emph{confirmation bias} \cite{Mercier2011}, the existence of separated cognitive modules %for different types of rules
\cite{Cosmides2008}, different framing processes % for descriptive and prescriptive rules 
\cite{Stenning2008,Kowalski2010a}, semantic and pragmatic factors influencing the reasoning task \cite{Stenning2008}, intervention of dual processing or heuristic-analytic models \cite{Evans2008}. The present paper presents yet another proposal, but its primary objective is rather to introduce an alternative view on the problem. Instead of focusing on the artificial, puzzle-like setting of selection tasks (which is problematic---respondents usually ask explicitly ``where is the trick?''), our investigation started from studying the mechanisms of construction of rule-like conceptual structures, generally abounding in human explicit knowledge (taxonomies, mereonomies, realization structures, etc.). In this line of research, Feldman \cite{Feldman2006} has shown that, by starting from some property language in which an observer describes objects and structure of the world, regularities emerging from the observations can be captured by a \textit{simplification} of an algebraic structure made of implication polynomials (related to Horn clauses). In particular, he proofs that species (e.g. bird as associated to properties like having wings, feathers, etc.) and trees of species (e.g. robins as sub-category of birds) can be seen as \textit{linear concepts}, i.e. structures that enable a simple algebraic decomposition. 

The idea that human cognition is fundamentally based on simplification mechanisms is an appealing one and has driven several research tracks in the last decades \cite{Chater1999,Chater2003,Dessalles2013}; in \textit{algorithmic information theory}, this has been powerfully synthesised by the expression ``\textit{understanding is compression}'' \cite{Chaitin2005}: the efficacy of a model, theory or other conceptual structures can be measured by how much they simplify our cognitive burden, how many observations they can summarise. If this is true, at qualitative level, the introduction of a new conceptual structure (like for instance a rule expressed through a logic conditional) requires at least to satisfy some compression property. Working on this intuition, this paper will:
\begin{itemize}\vspace{-5pt}
\item present an overview of the notion of \textit{supervenience} and identify it as a requirement for compression (sections 2.1, 2.2);
\item present a method relying on \textit{ontological dependence} to form a \textit{closure} of the antecedent of a logic conditional so that the consequent supervenes it (section 3.1);
\item confirm that taxonomies (as e.g. trees of species) and conceptual compositions (as e.g. species) are compressing structures, identifying two \textit{closure assumptions} (CA-I and CA-II) (sections 3.2, 3.3);
\item present and explain the performance of people in selection tasks through the lens of these closure assumptions (sections 4.1, 4.2).
\end{itemize}

\vspace{-10pt}
\section{Supervenience}
\vspace{-5pt}
Let us consider a class of objects $O$ that can be described with two properties, $a$ and $b$. Each property may hold or not, cases respectively identified with T and F. If we assume that there is no law that binds one property to the other, then we have 4 possible configurations that can be represented as a \emph{truth table}, or, in a similar spirit to e.g. \cite{Feldman2006}, as a \emph{Boolean lattice} ($\nn{a}$ stands for $a$ being F):
\vspace{-20pt}

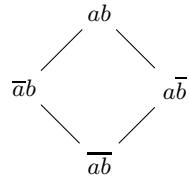
\begin{table}[h]	
	\begin{subtable}{.45\linewidth}
		\centering	
		\begin{tabular}{cc}
			$a$ &    $b$ \\ \hline
			T   &    T    \\  
			T   &    F    \\  
			F   &    T    \\   
			F   &    F       
		\end{tabular}
	\end{subtable}
	\begin{subtable}{.45\linewidth}
		\centering
		\begin{tikzpicture}[font = \sffamily, scale= 1]
		\path  (0,  0)  node (X)  {$ab$} 
		++(-1,-1)  node (Y)  {$\nn{a}b$}
		++(1, -1)  node (U)  {$\nn{ab}$}
		++(1,  1)  node (Z)  {$a\nn{b}$};
		\draw  (X) -- (Y);
		\draw  (X) -- (Z);
		\draw  (Y) -- (U) ;
		\draw  (Z) -- (U) ;  
		\end{tikzpicture}
	\end{subtable}
\end{table}

\vspace{-30pt}
\paragraph{The asymmetry of logic conditionals.}
\noindent Suppose now that the following rule holds:
$ a \rightarrow b $.
Interpreted as a \emph{material implication}, the logical conditional corresponds to the constraint $ \neg a \vee b $. Because the rule holds, we can remove the configurations for which it is false. Differently from assertions based on operators like $\vee$ and $\wedge$, a logic conditional introduces an \emph{asymmetry} on the lattice constructed with the possible configurations of $a$ and $b$:

\vspace{-15pt}
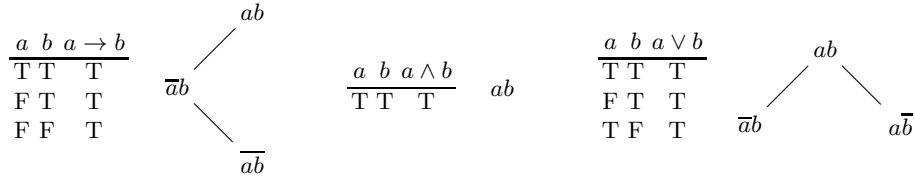
\begin{table}[h]
	\centering
	\hspace{-35pt}
	\begin{subtable}{.17\linewidth}
	\centering
	\begin{tabular}{ccc}
		$a$ &    $b$  & $ a \rightarrow b $ \\ \hline
		T &      T &      T \\ 
		F &      T &      T \\ 
		F &      F &      T 
	\end{tabular}
	\end{subtable}
	\begin{subtable}{.13\linewidth}
	\centering
	\begin{tikzpicture}[font = \sffamily, scale= 1]
	\path  (0,  0)  node (X)  {$ab$} 
	++(0, -2)  node (U)  {$\nn{ab}$}
	++(-1,  1)  node (Z)  {$\nn{a}b$};
	\draw  (X) -- (Z);
	\draw  (Z) -- (U) ;  
	\end{tikzpicture}
	\end{subtable}\qquad
	\begin{subtable}{.17\linewidth}
		\centering
		\begin{tabular}{ccc}
			$a$ &    $b$  & $ a \wedge b $ \\ \hline
			T &      T &      T 
		\end{tabular}
	\end{subtable}%
	\begin{subtable}{.05\linewidth}
		\centering
		\begin{tikzpicture}[font = \sffamily, scale= 1]
		\path  (0,  0)  node (X)  {$ab$};
		\end{tikzpicture}
	\end{subtable}\qquad%
	\begin{subtable}{.17\linewidth}
		\centering
		\begin{tabular}{ccc}
			$a$ &    $b$  & $ a \vee b $ \\ \hline
			T &      T &      T \\ 
			F &      T &      T \\ 
			T &      F &      T 
		\end{tabular}
	\end{subtable}%
	\begin{subtable}{.13\linewidth}
		\centering
		\begin{tikzpicture}[font = \sffamily, scale= 1]
		\path  (0,  0)  node (X)  {$ab$} 
		++(-1,-1)  node (Y)  {$\nn{a}b$}
		++(2,  0)  node (Z)  {$a\nn{b}$};
		\draw  (X) -- (Y);
		\draw  (X) -- (Z);
		\end{tikzpicture}
	\end{subtable}%

\end{table}
\vspace{-20pt}

% Natural sciences, for instance, approach reality depending on various factors, such as the dimensional scale in focus (e.g. particle physics vs astrophysics). This is because theories and accounts associated to the different \emph{levels} of reality are often so incompatible, that they may be seen as targeting different realities. In contrast, the position sustainining that all these levels can be reduced to one is usually called "reductionism". The motivation behind the introduction of supervenience lies therefore in attempting to furnish a framework compatible with the analysis and treatment of \emph{emergent} properties or phenomena, arising out of more \emph{fundamental} ones, but not \textit{reducible} to them.  But there are many other domains in which the notion of supervenience might be applied: computer scientists, for instance, don't look at computation considering the electrons performing it (information supervenes matter). 

\vspace{-15pt}
\subsection{Weak supervenience and determination}
In order to appreciate the sense of the asymmetry of logic conditionals, we investigated a more general asymmetric notion: \emph{supervenience}, introduced in modern philosophy in the attempt to capture the relation holding amongst different \emph{ontological levels} or \emph{strata}\footnote{As an example of \textit{ontological stratification}, consider how natural sciences specialize depending on the dimensional scale in focus (e.g. particle physics vs astrophysics). This is because theories and accounts associated to the different \emph{levels} of reality are often so incompatible, that they may be seen as targeting different realities.}, and so enabling the analysis and treatment of \emph{emergent} properties or phenomena arising out of more \emph{fundamental} ones, but not \textit{reducible} to them  \cite{Brown2009}.\footnote{Historically, \textit{supervenience} was introduced in support of the recognition of ``the existence of mental phenomena, and their non-identity with physical phenomena, while maintaining an authentically physicalist world view'' \cite{Brown2009}.} 

In the simplest form, ``we have supervenience when there could be no difference of one sort without differences of another sort'' \cite[p.~14]{Lewis1986}. This definition seems relatively simple; unfortunately, the literature exhibits a proliferation of non-equivalent formalizations\footnote{In addition to ``weak'' supervenience, we have ``strong'', ``global'' and many others, see e.g. \cite[p.~79]{Kim1993}, \cite{Stalnaker1996}.}, distinguished by several \emph{modal interactions}, distinct in number and placement of quantifiers over possible worlds or the type of necessity operators. An interesting exception, that we will use as a starting point, is the non-modal logic analysis given in \cite{Yoshimi2007}: 

\begin{definition}[Weak supervenience]A set of properties $B$ supervenes another set of properties $A$ if, 
	given any two entities $x$ and $y$, $x$ cannot differ from $y$ w.r.t. $B$ properties, without being different w.r.t. $A$ properties.
	$$B \;\textrm{supervenes}\;A \quad \equiv \quad \forall x, y : x \neq_B y \rightarrow x \neq_A y$$
\end{definition}

As we can see, supervenience decouples the concerns between a \emph{base} and a \emph{supervenient} sets of properties (the \emph{strata} or \textit{levels} we referred-to before), and identifies a directional relation between the two, defined by a constraint on \emph{differences}. Supervenience captures an abstract \emph{asymmetric covariation}: when it holds, a difference in the supervenient set let us expect a concurrent difference in the base set, but we do not define \emph{which} type of difference, nor \emph{why}.\footnote{See also Kim \cite[p.~167]{Kim1993} (italics mine):``[..] supervenience itself is not an explanatory relation. It is not a `deep' metaphysical relation; rather, it is a `surface' relation that reports a pattern of \emph{property covariation}, \emph{suggesting} the presence of an interesting \emph{dependence} relation that might explain it''. 
} Furthermore, despite what many believe, the `weak' formulation of supervenience is \emph{not} sufficient to be certain of the existence of a direct dependency between the two sets (e.g. that an object having a property in $A$ has also a property in $B$).  
% Here we provide a more formal treatment.
\noindent As observed in \cite{Yoshimi2007}, the \emph{contrapositive} of the formula of supervenience only identifies a \emph{determination} in terms of ``partial'' structural equalities: 

\begin{definition}[Determination]A set of properties $A$ determines another set of properties $B$ if, 
	given any two entities $x$ and $y$, when $x$ and $y$ are equal w.r.t. $A$ properties, they are equal w.r.t. $B$ properties.
	$$A \;\textrm{determines}\;B \quad \equiv \quad \forall x, y : x =_A y \rightarrow x =_B y$$
\end{definition}

%FALSE: it is a multiset injectivity Alternatively, we could collect the correlations of description of entities in $B$-terms and $A$-terms as a relation: $\rho: 2^A \rightarrow 2^B$. Denoting with $x_A$ a description of an entity $x$ through the set of properties $A$, we have $x_B = \rho(x_A)$, and then supervenience corresponds to the constraint: $$\forall x, y: x_B \ne y_B \rightarrow \rho(x_B) \ne \rho(y_B)$$
%which corresponds to $f$ be \emph{injective}.  

\vspace{-15pt}
\subsection{Supervenience and compression} Despite the ``weak'' formulation, supervenience can be seen as a requirement for \emph{compression}, when the \emph{base} set $A$ and the \emph{supervening} set $B$ count as providing symbols for \emph{encodings} of entities of a given domain $O$. 
Let us denote with $x_A = \rho_A(x)$ the description of an element $x$ through the set of properties $A$. Suppose we collect all co-occurrences of descriptions of all entities in $O$ in $A$-terms and in $B$-terms as instances of a relation $\rho_{AB} \subseteq 2^A \times 2^B$. In general this relation is not a function: two different objects $x$ and $y$ might exhibit equality w.r.t. $A$ ($x_A = y_A$) but not w.r.t to $B$ ($x_B \ne y_B$), and when this will happen an item in $2^A$ will relate to two items in $2^B$, so violating the right-unique property required for functional relations. If supervenience holds, however, equality w.r.t. $A$ will determine equality w.r.t. $B$, and so $\rho_{AB}$ becomes a function. In these conditions, there is a mapping between the encoding $A$ towards the encoding $B$:
$$ \rho_B(x) = \rho_{AB}(\rho_A(x)) $$
Stated differently, weak supervenience is required to \textit{re-encode} the description of the object based on $A$-properties  into a description in terms of $B$-properties. Imposing criteria of minimality on the $B$-encoding, the re-encoding becomes a \textit{compression} (generally a lossy one, as the mapping $\rho_{AB}$ might not be an isomorphism). % However, note that given a difference w.r.t $B$, supervenience grants us that there will a difference w.r.t $A$, but it does not indicate what this difference will be. 

\vspace{-3pt}
\section{Supervenience and logic conditionals}
At first sight, the expression of supervenience in terms of determination (Def.~2) seems to include the case of the implication expressed by a logic conditional, or at least to be related to.\footnote{Logic conditional is expressed between two individual properties, determination, as defined here, between sets of properties.} However, going through the possible configurations shown on the truth table of the conditional, when $b$ varies from T to F, $a$ may vary (first and third row) but it may also remain F (second and third row):

%\vspace{-20pt}
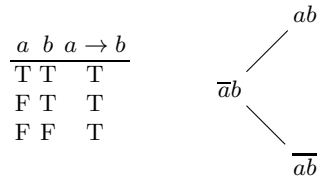
\begin{table}[ht]
\begin{subtable}{.45\linewidth}
	\flushright	
	\begin{tabular}{ccc}
		$a$ &    $b$  & $ a \rightarrow b $ \\ \hline
		T &      T &      T \\ 
		F &      T &      T \\ 
		F &      F &      T 
	\end{tabular}
\end{subtable}%
\begin{subtable}{.30\linewidth}
	\centering
	\begin{tikzpicture}[font = \sffamily, scale= 1]
	\path  (0,  0)  node (X)  {$ab$} 
	++(0, -2)  node (U)  {$\nn{ab}$}
	++(-1,  1)  node (Z)  {$\nn{a}b$};
	\draw  (X) -- (Z);
	\draw  (Z) -- (U) ;  
	\end{tikzpicture}
\end{subtable}
\vspace{-15pt}
\end{table}

\noindent In determination terms, this means that we can find two entities of this class which are equal with respect to $a$, but which are not equal in respect to $b$; therefore, \emph{supervenience is not} \emph{satisfied with a simple conditional}.

To repair this problem, we should consider a relation that instantiates that $a$ always varies when $b$ varies across the configurations. We discover that the resulting truth table is that of a \emph{bi-implication} (logical equivalence), introducing again a strong symmetry (actually \emph{replication}) amongst the two properties:
\vspace{-17pt}
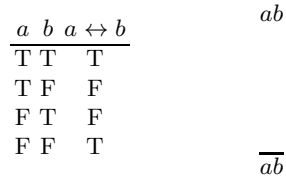
\begin{table}[!ht]
\begin{subtable}{.45\linewidth}
	\flushright
	\begin{tabular}{ccc}
		$a$ &    $b$  & $ a \leftrightarrow b $ \\ \hline
		T &      T &      T \\ 
		T &      F &      F \\ 
		F &      T &      F \\ 
		F &      F &      T 
	\end{tabular}
\end{subtable}\qquad\qquad%
\begin{subtable}{.30\linewidth}\quad
\begin{tikzpicture}[font = \sffamily, scale= 1]
   \path  (0,  0)  node (X)  {$ab$} 
        ++(0, -2)  node (U)  {$\nn{ab}$};
\end{tikzpicture}
\end{subtable}
\end{table}
\vspace{-17pt}

\noindent If this would be the only repair possibility, however, the introduction of  supervenience would bring about quite a slim advantage.

\subsection{Ontological dependence}
As we said above, weak supervenience specifies that there is a asymmetric relation between representations made with two sets of properties, but the two sets may be completely unrelated; for instance, the base set may be empty (\textit{free-floating} paradox). To avoid this case, Yoshimi \cite{Yoshimi2007} proposes to capture supervenience by adding to weak supervenience the constraint of \textit{ontological dependence}. %\footnote{An entity $a$ is ontologically dependent from an entity $b$, if $a$ cannot exists without $b$.} % \footnote{A relevant question is whether there are other ways to obtain supervenience.} 
%, thus realizing necessarily supervenience when satisfied. % % TO BE PROVED
In property terms, this would be written as:

\begin{definition}[Ontological dependence]A set of properties $B$ depends on another set $A$ only if all entities that exhibit a $B$-property, exhibit as well a $A$-property:
$$B \;\textrm{depends on}\;A \quad \equiv \forall x, \beta \in B: \beta(x) \rightarrow \exists \alpha \in A: \alpha(x)$$
\end{definition}
This property binds the manifestation of each property in $B$ to the manifestation of at least a property in $A$. Using the contrapositive, it can be also read as a sort of \emph{closure}: if an object does not manifest any property in $A$, then it does not manifest any property in $B$. 

%$$ \forall x, \beta \in B: [\beta(x) \rightarrow [\exists \alpha \in A]: \alpha(x) ]$$
%$$ \forall x, \beta \in B: [ [\neg \exists \alpha \in A: \alpha(x)] \rightarrow \neg \beta(x) ]$$
%$$ \forall x, \beta \in B: [ [\forall \alpha \in A: \neg \alpha(x)] \rightarrow \neg \beta(x) ]$$

% \vspace{-10pt}
% \begin{table}[ht]
% 	\centering
%	\begin{tabular}{cccccc}
%	$a^{F}$ & $a^{T}$	& $a^*$ &	$a$ &    $b$  & $ a \rightarrow b $ \\ \hline
%	F &	T    & X       & T      &  T  &      T \\
%	F &	T    & T       & F      &  T  &      T \\
%	F &	T    & F       & F      &  F  &      T \\
%	\end{tabular}
%\end{table}
%\vspace{-10pt}

\subsubsection{Logic conditionals and ontological dependence}
In the case of a conditional, restricting $A$ to a single property $a$ and $B$ to a single property $b$, ontological dependence is equivalent to the inverse of the conditional, and therefore instantiates the bi-implication case seen above, which we know to satisfy supervenience. However, the definition given above suggests an alternative path. Let us consider an additional property $a^*$ which is true when $b$ is true and $a$ is false (X means any truth value is possible):
\vspace{-10pt}
\begin{table}[ht]
	\centering
	\begin{tabular}{cccc}
		$a^*$ &	$a$ &    $b$  & $ a \rightarrow b $ \\ \hline
		X       & T      &  T  &      T \\
		T       & F      &  T  &      T \\
		F       & F      &  F  &      T 
	\end{tabular}
\end{table}
\vspace{-10pt}

\noindent With this simple addition, the set of properties $B = \{b\}$ supervenes the set $A^* = \{a, a^*\}$, without modifying the constraint set by the initial logic conditional. This result can be generalized: \textit{the consequent of a conditional supervenes the antecedent, if adequately closed through ontological dependence}. In practice, the closure set consists of all sufficient conditions determining the consequent. In this case, because the one provided was not sufficient to produce all manifestations of the consequent, a new condition has been introduced. 

\subsection{Supervenience in subsumption}

% In this way, generalization, abstraction (\emph{subsumption} in predicate form) of $a$ to $b$ can be interpreted in terms of supervenience. 

Let us consider a typical use of conditionals at class level, i.e. in subsumption rules as ``all dogs are animals'': 
$$\forall x: \mathit{Dog}(x) \rightarrow \mathit{Animal}(x)$$
The closure constraint identified above states that supervenience holds (i.e. the predicate $\mathit{Animal}$ can compress), only if we consider all the sufficient conditions for an entity $x$ to be animal, i.e. $x$ belonging to any of the subclasses of animal. This is intrinsic to the very idea of class: \textit{it is not possible that being animal is true without having any of its known subclasses true}:
$$\neg \exists x: \mathit{Animal}(x) \wedge \neg \mathit{Dog}(x) \wedge \neg \mathit{Cat}(x) \wedge \ldots $$
This closure assumption (hereby denoted with CA-I) is an intuitive principle at the base of all taxonomical relations.\footnote{For simplicity, we are overlooking here the aspects related to hierarchization, e.g. exclusive disjunction of subclasses at a given level of depth.} Here, more in general, we have shown that the presence in the knowledge of a logic conditional \textit{joint with} the associated CA-I implies that the consequent compresses the closure of the antecedent: \begin{itemize}
\item when the consequent is false, all possible antecedents are false as well;
\item when the consequent is true, at least one antecedent is true.
\end{itemize} 
In these conditions, \textit{modus tollens} can remove all antecedents in the closure at once, but also the conditional can be disproven just by finding one antecedent in the closure which is true when the consequent is false.

\subsection{Supervenience in conceptual composition}
The asymmetry of the conditional may be seen in the opposite sense, reading the absence of a property as a property, an then
%\vspace{-10pt}
%
%\begin{table}[ht]
%	\centering
%	\begin{tabular}{cccccc}
%		$a^*$ &	$a$ &    $b$  & $ a \rightarrow b $ & $\neg{a}$ & $\neg{b}$ \\ \hline
%		X       & T      &  T  &      T  & F & F \\
%		T       & F      &  T  &      T  & T & F \\
%		F       & F      &  F  &      T  & T & T 
%	\end{tabular}
%\end{table}
%\vspace{-10pt}
considering the \emph{contrapositive} of the initial conditional ($\neg b \rightarrow \neg a$). As in the previous case, in order to guarantee supervenience of the consequent $\neg a$, we need to introduce an adequate $\neg{b^*}$ for closure related to ontological dependence:

%\vspace{-5pt}
%\begin{table}[ht]
%	\centering
%	\begin{tabular}{ccccccc}
%	$a^*$ &	$a$ &    $b$  & $ a \rightarrow b $ & $\nn{a}$ & $\nn{b}$ & $\nn{b^*}$ \\ \hline
%X       & T      &  T  &      T  & F & F & F \\
%T       & F      &  T  &      T  & T & F & T \\
%F       & F      &  F  &      T  & T & T & X 
%	\end{tabular}
%\end{table}
%\vspace{-5pt}

\vspace{-15pt}
\begin{table}[ht]
	\centering
	\begin{tabular}{ccccccc}
		$a^*$ &	$a$ &    $b$  & $ a \rightarrow b $ & $\neg{a}$ & $\neg{b}$ & $\neg{b^*}$ \\ \hline
		X       & T      &  T  &      T  & F & F & F \\
		T       & F      &  T  &      T  & T & F & T \\
		F       & F      &  F  &      T  & T & T & X 
	\end{tabular}
\end{table}
\vspace{-20pt}

\noindent In this new configuration, we have that $\neg A = \{\neg{a}\}$  supervenes the set $\neg B^{*} = \{\neg{b}, \neg{b^*}\}$. Thus, the negation of the antecedent of a conditional supervenes the negation of the consequent, if adequately closed through ontological dependence.
Interestingly, this closure can be related to \textit{conceptual compositional} structures (including mereonomies, realization structures, causal mechanisms, etc.). Consider for instance the following rule:
$$\forall x: \mathit{Dog}(x) \rightarrow \mathit{hasTail}(x)$$
% $$\forall x: \mathit{Dog}(x) \rightarrow \exists y: \mathit{Tail}(y) \wedge \mathit{Has}(x, y)$$
%; therefore the composition of $b$ with $a$ (or the aggregation of $a$ to $b$) can be specified in terms of supervenience as well. For instance, all dogs have a tail\footnote{Clearly, in normal conditions. We are neglecting for the moment the inclusion of defaults, that would eventually bring to the Aristotelian distinction between essential and accidental properties.}:
Although we could in principle consider dogs as a sub-class of the greater class of entities with a tail, the contrapositive formulation captures a different type of closure: % A tail is \emph{part-of} a dog (i.e. having a tail is a necessary property of a dog), if the property of \emph{not} being a dog supervenes that of \emph{not} having a tail and any other part of the dog. 
$$\neg \exists x: \neg \mathit{Dog}(x) \wedge \mathit{hasTail}(x) \wedge \mathit{hasFur}(x) \wedge \dots$$
\noindent In words, \textit{one entity cannot have all the properties associated to certain class without being of that class}. %\footnote{A similar formulation would be ``one entity cannot have all the parts associated to a certain whole without being that whole''. The specificity of mereology in respect to property aggregation remains to be investigated.} %(If this happens, we could trigger an abduction process to identify the additional discriminating property). 
This closure assumption (hereby denoted with CA-II) is at the base of the composition/aggregation of properties as higher-order properties (wholes, higher-order actions, cause-effects, etc.). In more general terms, the assertion of a logic conditional with the associated CA-II implies that the antecedent compresses the closure of the consequent: 
\vspace{-3pt}
\begin{itemize}
	\item when the antecedent is true, all possible consequents are true as well;
	\item when the antecedent is false, at least one consequent is false.
\end{itemize}
\vspace{-3pt} 
In these conditions, \textit{modus ponens} can produce all consequents in the closure at once, but also the conditional can be disproven just by finding one consequent in the closure which is false when the antecedent is true.

\vspace{-5pt}
\section{Reviewing Wason's selection tasks}
\vspace{-5pt}
Exploiting the framework just presented, we will propose an alternative explanation of human performance in selection tasks, a famous class of behavioural psychology experiments introduced by Wason at the end of the 1960s \cite{Wason1968}.

\vspace{-10pt}
\subsubsection*{Selection tasks}

Given a simple rule (usually in the conditional form), respondents are asked to \textit{select}, amongst few instances, the ones which are relevant to check whether the rule applies. % This short paper takes a sort of neutral perspective: it accepts these biases as evolutionary justified, and then as a proper (and not biased) form of individual inference;  
% Represented in a logic form, the problem is 
For instance:
\begin{example}
	It has been hypothesized that if a person has Ebbinghaus disease, he is forgetful. You have four patients in front of you: A is not forgetful, B has the Ebbinghaus disease, C is forgetful, and D does not have the Ebbinghaus disease. Which patients must you analyse to check whether the rule holds?
\end{example}

\noindent In classic logic, when a rule $p \rightarrow q$ holds, also the contrapositive $\neg q \rightarrow \neg p$ holds. Therefore to check whether a rule holds, you must check:
\begin{enumerate}
	\item whether the individuals that exhibit $p$ exhibit $q$  as well, and 
	\item whether the individuals that don't exhibit $q$, don't exhibit $p$ either. 
\end{enumerate}
In the previous example, these answers are respectively B and A. 

Unexpectedly, evidence shows that humans in tendency perform only the first of these checks, but not the second. A number of respondents also select the logically wrong choice $q$, that could be associated to a \textit{biconditional} reading of the rule. Similar results have been observed with selection tasks based on common-sense and expert domains, and in experiments with layman and experts (including mathematicians). However, experiments on selection tasks have also shown clear exceptions to these results when tasks build upon \emph{deontic rules}, i.e. rules about norms of conduct. Consider for instance the following task:

\begin{example}
	In your country, a person is not allowed to drink alcohol before the age of 18. You see four people in a pub: A is enjoying his beer, B is drinking an orange juice, C is at least 40 years old, and D is no older than 16 years. Which people must you investigate to check whether the rule is applied?
\end{example}
In this case, the great majority of respondents select A and D, the logically correct answers.

\vspace{-5pt}
\subsection{Explaining selection tasks via closure assumptions}
Many hypotheses have been proposed in the literature to explain these phenomena (see section 1). We construct here yet another explanation, relying on the closure assumptions identified above. Let us review the previous selection tasks and two additional scenarios (a subsumption rule and the first selection task proposed by Wason):
\begin{itemize}
	\item[(1)] \textit{If a person has Ebbinghaus disease, then he is forgetful.}
	\begin{itemize} \vspace{1pt}
		\item CA-I: one cannot be forgetful, without having the Ebbinghaus disease (or any other known cause of being forgetful). \vspace{1pt}
		\item CA-II: one cannot be forgetful (and any other known effect of the Ebbinghaus disease) without having the Ebbinghaus disease.
	\end{itemize}
\end{itemize}
When presented to a \textit{causal rule} of the type \textit{disease-producing-symptom}, people are suggested to apply CA-II, but do not naturally apply CA-I. In effect, there may be always unknown causes for becoming forgetful, thus, the property of being forgetful cannot ascend to the status of class. On the other hand, the framing given by the experiment invite to think that being forgetful is the only discriminating symptom of the Ebbinghaus disease. 

%\begin{itemize}
%	\item[(2)] \emph{If you drink alcohol, then you are older than 18 years old.}\footnote{The rule is here purged from deontic modalities. A modal version following the institutional causation would be e.g. ``\emph{If you are older than 18 years old, then you are allowed to drink alcohol.}'', which, rephrased in teleological form, provides a rule similar to the indicative one considered above: ``\emph{In order to drink alcohol, you have to be older than 18 years old.``}}
%	\begin{itemize}
%		\item CA-I: one cannot be older than 18 years old, without drinking alcohol and with drinking alcohol (there are clearly adults of both types).
%		\item CA-II: one cannot be older than 18 years old (and any other conditions), without drinking alcohol.
%	\end{itemize}
%\end{itemize}
%In this case, CA-I is always satisfied because it includes the conjunction of a predicate and of its contrary. Being older than 18 years old ascends to the \textit{class} status. Also CA-II may be in principle satisfied (we usually know in which conditions something is allowed or prohibited).

\begin{itemize}
	\item[(2)] \emph{If you are older than 18 years old, then you are allowed to drink alcohol.}\footnote{The proposed modal version follows the institutional causation. Rephrased in teleological form, it becomes: ``\emph{In order to drink alcohol, you have to be older than 18 years old.}'', that can be related to the indicative form: ``\emph{If you drink alcohol, then you are older than 18 years old.}``}
	\begin{itemize} \vspace{1pt}
		\item CA-I: one cannot be allowed to drink alcohol, without being older than 18 years old (or any other known requirement for drinking). \vspace{1pt}
		\item CA-II: one cannot be allowed to drink alcohol (and any other known condition associated to being older than 18 years old), without being older than 18 years old.
	\end{itemize}
\end{itemize}
\vspace{-2pt}
In this case, CA-I and CA-II can be assumed to hold because the very communicative purpose of norms is to provide knowledge about both the qualifying conditions and of their normative effects in a certain social domain.\footnote{From this consideration we can trace the hypothesis that, when confronted with complex normative structures, people will turn to a conservative, ``causal-rule'' type of interpretation.}
\vspace{-2pt}
\begin{itemize}
	\item[(3)]\textit{ If an entity is a dog, then it is an animal.}  
	\begin{itemize} \vspace{1pt}
		\item CA-I: an entity cannot be an animal, without being a dog, or belonging to any other subclass of the animal kingdom. \vspace{1pt}
		\item CA-II: an entity cannot be an animal and all other known properties discriminating a dog entity, without being a dog. 
	\end{itemize}
\end{itemize}
\vspace{-2pt}
In this case, both CAs apply, otherwise the concepts of dog and of animal wouldn't be functioning properly.
\vspace{-2pt}
\begin{itemize}
	\item[(4)]\textit{If there is ``D'' on one side of a card, then there is ``3'' on the other side.}  
	\begin{itemize} \vspace{1pt}
		\item CA-I: a card cannot have ``3'' on one side, without having ``D'' on the other side (or any known other symbol mapping from ``3''). \vspace{1pt} 
		\item CA-II: a card cannot have ``3'' on one side, % (and any other known symbol mapping from ``D''), 
		without having ``D'' on the other side.
	\end{itemize}
\end{itemize}
\vspace{-2pt}
Here CA-II is certainly true: for the 2-sides structure of cards, only one association is possible. Viceversa, CA-I is not acceptable in general, because the other configurations mapping to ``3'' are not known. 

\vspace{-9pt}
\subsubsection{Cognition-as-compression hypothesis} The previous analysis suggests a way to predict which behaviour will be selected by people depending on the conceptual structure associated to the proposed selection task. The general scenario (usually associated to \textit{descriptive rules}) relates to cases in which only CA-II applies; supposing the task is in the form $p \rightarrow q$, people will mostly select items exhibiting $p$ to test whether $q$ is the case. The exceptional scenario, but theoretically correct one (usually associated to \textit{prescriptive rules}), relates to cases in which both CA-I and CA-II apply; here, people will test $\neg q$ as well. 

This behaviour can be explained by assuming that people interpret conditionals in different ways depending on the attributed \textit{compression capacity}, which in turn depends on their domain conceptualization (but not on the descriptive/prescriptive nature of the rule). 

\pagebreak
\noindent Stated differently, in selection tasks, they don't test the CAs, but they test the conditional's compression capacity enabled by the CAs:
\vspace{-3pt}
\begin{itemize}
	\item when only CA-II is deemed to apply, people focus on the antecedent: \textit{does the antedecent effectively compress?} To confirm this, one has to check whether, when the antecedent is present, all consequents in the closure are present as well (i.e. that $q$ holds on $p$), see section 3.2. 
	\item when both CA-I and CA-II are deemed to apply,  people concentrate on both antedecent and consequent. In addition to the previous test, \textit{does the consequent effectively compress}? To answer this question, if the consequent is absent, one has to check whether all antecedents in the closure are absent as well (i.e. that $\neg p$ holds on $\neg q$), see section 3.3.
\end{itemize}
\vspace{-3pt}
Interestingly, this account gives also some insights on why people might select $q$: the \textit{biconditional reading} corresponds to force supervenience (compressibility) on the conditional without looking at closure assumptions (see beginning section 3); but also on why the conditional might be deemed \textit{irrelevant} (e.g. as in Wason's \textit{defective truth table}): when only CA-II applies, and the antecedent is false, the compression mechanism is not activated.

\vspace{-6pt}
\section*{Conclusion}
\vspace{-3pt}
The intuition starting our investigation was that supervenience might act as a common denominator for ordering relations creating \textit{ontological stratifications}: \emph{taxonomical} structures (ordered by the subsumption),  \emph{mereological} structures (ordered by part-whole relations), \emph{supervenience} structures (ordered by asymmetric dependencies), \emph{realization} structures (ordered by functional relations), \emph{nomological} structures, etc. \cite{Brown2009}.
This paper introduces a crucial argument in support to this intuition: any meaningful abstraction requires compression, and supervenience counts as a necessary requirement for compression. We have then analysed through this lens subsumption and conceptual composition, in the sense of general aggregation of properties (relevant to mereological structures, realization structures and causal mechanisms), identifying two \textit{closure assumptions} enabling supervenience on logic conditionals. As an unexpected by-product, we obtained an alternative explanation of human performance in selection tasks.

By reframing the reasoning process activated by selection tasks in terms of evaluating the compression capacity of the proposed rules (taken as prototypical example of knowledge constructs) rather than testing their logic validity, our theory supports a positive view on human cognition, rather than a negative one (as suggested by words like \textit{bias}). 
More concretely, it shows that the distinction between general and exceptional performance is not caused by the content in itself (of descriptive or of prescriptive nature), but by the closure assumptions through which this is processed. This is compatible with other hypotheses insisting on contextual aspects: experimental framing can indeed modify such closure assumptions for the observers, but clearly also their personal knowledge and dispositions might play a role (see e.g. \cite{Counihan2008}: unschooled subjects commonly refuse to reason with given premises or provide their own premises as a basis for reasoning).

Going beyond the selection task literature, in recent times, \textit{indicative conditionals} have attracted a renewed interested, reinvigorating the debate around Adams' thesis: \textit{the acceptability of an indicative conditional sentence goes by the conditional probability of its consequent given its antecedent.} Empirical observations \cite{Douven2010} show that this principle is not descriptively correct, and several authors started working on identifying additional conditions of dependence (see e.g. \cite{VanRooij2019}). In future work, we plan to evaluate the proposal advanced in this paper w.r.t. these other contributions.

\end{document}